%% file: main.tex
\documentclass[dvipsnames,format=sigconf ]{acmart}

\usepackage{graphicx} 
\usepackage{subcaption} 
\usepackage{xspace}
\graphicspath{{./images/}} 
\input{notations}

\usepackage{xcolor}

\usepackage{hyperref}

\copyrightyear{2023}
\acmYear{2023}
\setcopyright{rightsretained}
\acmConference[GECCO '23 Companion]{Genetic and Evolutionary Computation Conference Companion}{July 15--19, 2023}{Lisbon, Portugal}
\acmBooktitle{Genetic and Evolutionary Computation Conference Companion (GECCO '23 Companion), July 15--19, 2023, Lisbon, Portugal}\acmDOI{10.1145/3583133.3590625}
\acmISBN{979-8-4007-0120-7/23/07}

\begin{document}

\title{Quality-Diversity Optimisation on a Physical Robot Through Dynamics-Aware and Reset-Free Learning}

\author{Sim\'on C. Smith}
\email{s.smith-bize@imperial.ac.uk}
\authornotemark[1]
\authornotemark[2]
\affiliation{%
  \institution{}
  \city{}
  \country{}
}
\thanks{$^*$ School of Computing, Engineering and The Built Environment, Edinburgh Napier University, UK.\\
$^\dag$Department of Computing, Imperial College London, UK.\\
This work was partially supported by the Engineering and Physical Sciences Research Council [grant number EP/V006673/1]
.}

\author{Bryan Lim}
\email{bryan.lim16@imperial.ac.uk}
\authornotemark[2]
\affiliation{%
  \institution{}
  \city{}
  \country{}
}

\author{Hannah Janmohamed}
\email{hannah.janmohamed21@imperial.ac.uk}
\authornotemark[2]
\affiliation{%
  \institution{}
  \city{}
  \country{}
}

\author{Antoine Cully}
\email{a.cully@imperial.ac.uk}
\authornotemark[2]
\affiliation{%
  \institution{}
  \city{}
  \country{}
}

\renewcommand{\shortauthors}{Smith, et al.}

\begin{abstract}

Learning algorithms, like Quality-Diversity (\qd), can be used to acquire repertoires of diverse robotics skills. This learning is commonly done via computer simulation due to the large number of evaluations required. However, training in a virtual environment generates a gap between simulation and reality. Here, we build upon the Reset-Free QD (\rfqd) algorithm to learn controllers directly on a physical robot. This method uses a dynamics model, learned from interactions between the robot and the environment, to predict the robot's behaviour and improve sample efficiency. A behaviour selection policy filters out uninteresting or unsafe policies predicted by the model. \rfqd also includes a recovery policy that returns the robot to a safe zone when it has walked outside of it, allowing continuous learning. We demonstrate that our method enables a physical quadruped robot to learn a repertoire of behaviours in two hours without human supervision. We successfully test the solution repertoire using a maze navigation task. Finally, we compare our approach to the \mapelites algorithm. We show that dynamics awareness and a recovery policy are required for training on a physical robot for optimal archive generation.
Video available at https://youtu.be/BgGNvIsRh7Q
\end{abstract}

\ccsdesc[500]{Computer systems organization~Evolutionary robotics}
\ccsdesc[500]{Theory of computation~Evolutionary algorithms}

\keywords{Quality-Diversity, Robotics, Real-time optimisation}

\begin{teaserfigure}
\centering
\includegraphics[width=.9\linewidth]{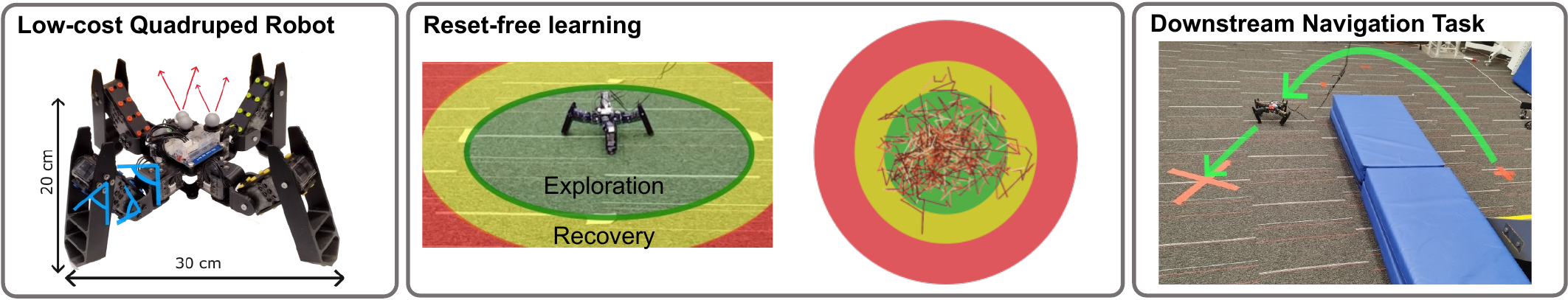}
\vspace{-8pt}\caption{
    The Qutee robot. A 12 DoF (blue angles) quadruped. Global coordinates given by a motion capture system (red arrows). 
    Middle: the trace of 2 hours of reset-free training within the exploration (green) and recovery (yellow) zones.
    Right: the navigation task used to test the learned solutions.
    All training was done on the physical robot, without any simulators. 
}
\label{fig:featurefig}
\end{teaserfigure}

\maketitle

\vspace{-10pt}
\section{Introduction}


Quality-Diversity (\qd) algorithms are techniques for optimisation that, among other, have been used for learning robotics controllers deployed in the real world. 
These algorithms find extensive collections of both high-performing and diverse solutions,
which is advantageous for downstream applications \cite{nature, hbr}. 
However, to the best of our knowledge, all existing \qd algorithms applied in robotics use computer simulations to evaluate controller candidates.
While this approach affords evaluation of thousands of solutions, it also comes with various disadvantages.
For example, simulators require accurate modelling of the physical properties and dynamics of the environment.
Though a great variety of advanced simulators have been developed
, it is still difficult to model the real world with high fidelity.
Furthermore, rigid-body simulation often assumes a controlled environment and a robot with accurate sensors and actuators.
Meanwhile, in the real world, sensors and actuators are often prone to noise and damage. 
Some methods, such as Kalman filters and high-frequency closed-loop control, attempt to mitigate the effects of stochastic environments.
However, the controllers learned in simulation are not always robust when transferred directly to physical robots. 
Consequently, \qd approaches in robotics usually encompass further adaptation mechanisms to bridge the gap between simulation and reality \cite{nature, hbr, daqd}
which require additional training and fine-tuning.
In this work, we use the Reset-free Quality-Diversity (\rfqd) algorithm \cite{rfqd} to learn repertoires of controllers directly on a physical quadruped robot without human intervention.
\rfqd learns a dynamics model to predict the behaviour of controllers.
Based on these predictions, \rfqd applies a behaviour selection policy to prioritise controllers according to predefined safety and exploration heuristics to be executed in the environment.
Accordingly, controllers that are predicted to be unsafe or uninteresting are filtered out, leading to better sample efficiency, reduced training time and decreased risk of damage.
Furthermore, given environmental information, \rfqd ensures that training always takes place in a safe zone so that continuous learning can take place, without the need of human supervision or manual resets.

Our results show that \rfqd can successfully generate an archive of behaviours without simulation in two hours.
We compare \rfqd with baselines to show that the recovery policy and behaviour prioritisation are both essential components for achieving high-performing and diverse repertoires.
Finally, we test the generated repertoires on a navigation task. 

\vspace{-10pt}
\section{Related Work}

\textbf{Quality-Diversity in Robotics}. Quality-Diversity (\qd) form a subset of evolutionary algorithms that generate many diverse and high-performing solutions. 
To achieve this, \qd algorithms characterise solutions by their fitness, an objective function, and by a vector known as the \textit{behaviour descriptor} (\bd) \cite{qdunifying}.
The \bd is used to quantify the difference in the behaviour of solutions.

In general, \qd algorithms are initialised with a set of random solutions
stored in an archive.
At each iteration, solutions are selected from the archive and mutated to generate offspring.
The offspring are evaluated and considered for addition to the archive, based on their fitness and \bd. 
The precise addition mechanism differs according to the \qd algorithm.
For example, \mapelites \cite{mapelites} employs a grid-based archive, achieved by tessellating the \bd space into several fixed-size bins.
A solution is added to the grid if the corresponding cell is empty or its fitness is higher than an existing solution in the same cell, discarding the previous solution.
Alternatively, other \qd algorithms use an unstructured archive \cite{aurora, daqd} in which a solution is added if the distance to its k-nearest neighbours in the descriptor space exceeds a predefined threshold.

QD algorithms are used in robotics to learn a collection of controllers.
The diversity of skills learned is advantageous for use in downstream applications such as damage recovery or planning for long horizon tasks~\cite{hbr, rte, daqd}.
Existing methods first learn skills in simulation, and only after learning they use them on the physical robot.
In our work, we apply the QD algorithm directly on the physical robot, with no simulator and prior experience.


\textbf{End-to-end Learning on Physical Robots}. Data-driven and learning based methods are able to learn very complex skills~\cite{akkaya2019solving}.
However, they require a large number of samples or trials, making them infeasible for direct implementation on physical robots.
Learning directly on physical robots rely on model-based RL for locomotion~\cite{dataefificentrl, daydreamer} and manipulation~\cite{nagabandi, solar, daydreamer}.
Our work demonstrates learning directly on a physical robot is also possible for evolutionary-based algorithms such as \qd algorithms.



\section{Methods}


\textbf{Learning a Dynamics Model}. Following DA-QD~\cite{daqd}, a forward model $p(\boldsymbol s_{t+1}|\boldsymbol s_t,\boldsymbol a_t)$ that predicts the next state $\boldsymbol s_{t+1}$ given the current state $\boldsymbol s_t$ and action $\boldsymbol a_t$, is learnt based on data collected from interaction in the environment. 
The model is implemented as a deep neural network where the delta of the next state $(\Delta \boldsymbol s_{t+1} = \boldsymbol s_{t+1}- \boldsymbol s_{t})$ is learned. 
Similar to DA-QD and RF-QD, we learn an ensemble of probabilistic models to minimise both aleatoric and epistemic uncertainty.
The disagreement between the models in the ensemble is inferred using their distributions, allowing an estimation of the epistemic uncertainty, which can be used to prioritise new controllers.
The model is trained via self-supervised learning using gradient descent to maximise the log-likelihood of transitions sampled from the replay buffer.
The dynamics model is used to perform rollouts of controllers, called recursively for the defined length of the execution of a controller.

\textbf{Performing QD in imagination}. 
\rfqd uses the dynamics model to predict a robot's trajectory, fitness and behavioural descriptor for a given controller.
Solutions can be evaluated without requiring any real-world interaction.
Thus, the \qd loop of selection, mutation, evaluation and archive addition is performed using imagined rollouts and maintained in an separated archive.
Solutions from this imagined archive are selected to be executed in the real world.
These physical executions are added to the main archive and added to the reply buffer for further training of the dynamics model.
Performing evaluations in imagination allows better data efficiency as solutions that are not promising will be sieved out and not executed.

\textbf{Behaviour Selection Policy}. BSP ensures that the robot remains in a safe state while learning new skills and interacting with the environment. 
The BSP determines the solutions from the imagined archive to be executed 
provided that the user defines a safety signal.
In locomotion, BSP comprise exploration zones (a set of safe states) and recovery zones (a set of unsafe states) $\Omega$.
The relative safety of the robot in state $\boldsymbol s$ is measured by an exploration parameter $\epsilon(\boldsymbol s)$, calculated as the distance between $\boldsymbol s$ and the nearest unsafe state $\boldsymbol \omega \in \Omega$ normalised by the maximum distance between any previous state and $\boldsymbol \omega$:
\vspace{-1mm}
\begin{align}
    \epsilon(\boldsymbol s) = \frac{dist(\boldsymbol s, \boldsymbol \omega) - \beta}{\max_{\boldsymbol s_i} dist(\boldsymbol s_i, \boldsymbol \omega) - \beta}.
\label{eq:epsilon}
\end{align}
Here, $\beta$ ensures a buffer space when the robot returns to the exploration zone.

New controllers $\boldsymbol s'$ are safe if they are expected to keep the robot in the exploration zone, i.e. $\epsilon(\boldsymbol s')>0$.
Using this safety metric, RF-QD filters the solutions in the imagined archive to obtain a subset of solutions expected to be safe.
RF-QD prioritise solutions in this safe set by safety, novelty or disagreement between models.
In our experiments, we use novelty prioritisation to bias exploration in the \bd space.

\textbf{Recovery Policy}. A recovery policy is used for safety if the robot leaves the exploration zone.
While in the recovery zone, the policy selects behaviours in the archive that returns the robot to the exploration zone.
In the case of leaving the recovery and the exploration zone the learning is stopped. 

\textbf{Controllers Update}. We extend \rfqd to include an update strategy for controllers used during recovery. The executions of these controllers are used to update their fitness and \bd.
Given a controller with fitness $f$ and behavioural descriptor $\mathbf b$, the new values are given by:
\vspace{-2mm}
\begin{equation*}
f =(1-\alpha)f + \alpha f',~\mathbf b = (1-\alpha)\mathbf b + \alpha \mathbf b',
\end{equation*}
where $f'$ and $\mathbf b'$ are the fitness and behaviour descriptor values observed in the new execution, with $\alpha \in [0, 1]$. This updated controller is reevaluated for addition or removal from the archive.

\textbf{Physical Implementation}.
We use the Qutee robot (Fig.\ref{fig:featurefig}) which was designed, 3D printed and assembled by researchers at the Adaptive and Intelligent Robotics Lab at Imperial College.
One cable is attached to the robot for power and another for data transmission to a computer.

The action space of the robot corresponds to the angle commands sent to each of these DoF.
We restrict the range of movement of each DoF to avoid collisions between the legs and the body of the robot.
We restrict the range of the hip actuators in the horizontal plane to reduce the chances of turning upside down.

Exterioception is provided by a motion capture system.
The system returns the position, velocity, angle and angular velocity of the centre of mass of the robot. We transform the world coordinates to positions relative to the robot's initial location. 
These coordinates are used as part of the robot's state space, along with the position and velocities of each joint.


We define four main zones for exploration and recovery as visualised in Fig.~\ref{fig:featurefig}. 
The robot is initially placed inside the green exploration zone.
In this zone, the robot executes new controllers.
Based on their fitness and \bd, controllers are evaluated to be added to the archive following the usual QD rules. 
When executing a controller, if the robot exits the exploration zone, it enters the yellow recovery zone.
From this zone, the recovery policy is activated. The policy takes into account the $\beta$ (Eq.~\ref{eq:epsilon}) buffer before returning to the exploratory mode.
The training is stopped if the robot leaves the recovery zone (red in Fig.~\ref{fig:featurefig}).

\vspace{-2mm}
\section{Experiments and Results}

\subsection{Experiments and Ablations}
\textbf{Training}. We use an omnidirectional walking task to generate an archive of controllers. In this task, the robot learns walking policies to move in the horizontal plane~\cite{nature}. The \bd is the final position of the robot with respect to the initial position of the behaviour.
The fitness is the negative error between the horizontal rotational angle of the robot in the final position and the required angle to arrive following a predetermined arc from the initial position. The genotype is a vector of size 24.
This vector represents the parameters of an open-loop sinusoidal controller. The movement of each joint is parameterised by the amplitude, the phase and the duty cycle. We use the same parameters for the foot and knee of each leg.

We assume that the robot has a 2-D map of the exploration and recovery zone, represented as concentric circles of radii $0.5$m and $0.75$m respectively, with $\beta = 0.3$.
The execution of a controller last $5$s.
We set $\alpha=0.8$. 
To generate offspring, we use the ISO+LineDD mutator~\cite{10.1145/3205455.3205602}. We use an unstructured archive to store the solutions and define novelty and gradient-contextual constraints for the behaviour selection policy. For the dynamics model probabilistic ensemble, we use 4 neural networks with two hidden layers of $500$ neurons each.
We store the action-state pairs in a replay buffer and train the dynamics model after 10 evaluations.

Each experiment runs for $2$ hours or until the robot leaves the recovery zone. RF-QD includes an initialisation phase where 10 random controllers are generated, executed and added to the archive. Among all the experiments we ran, this initialisation failed only once due to the robot moving out of the recovery zone.

\textbf{Comparison to other algorithms}. We compare RF-QD to two variants.
\emph{RF-QD without dynamics awareness}, we remove the dynamics model.
Thus, no training or behaviour selection policy is possible. Removing the training phase allows more evaluations during the 2 hours of the experiment.
\emph{RF-QD without recovery function}, the recovery function is removed.
This version still has the dynamics model so the behaviour selection policy still bias the robot to remain in the exploration zone.
We compare the algorithms to MAP-Elites, which is equivalent to \rfqd without a dynamics model and no recovery policy.
All the meta-parameters remain the same between the four algorithms.
We trained each version 4 times and tested the solutions on navigation task 5 times each.

\textbf{Planning on a maze}. For testing, we use a maze task in which the robot must navigate to a goal while avoiding obstacles (Fig.~\ref{fig:featurefig}).
The location of the goal and obstacles are known to the robot.
We use RTE (Reset-free Trial-and-Error) algorithm~\cite{rte} with an A$^*$ planning algorithm to select the best solution from the archive.
We count a test as successful if the robot reaches the goal within a range of $5$cm, and unsuccessful if the robot executes more than 100 actions without reaching the goal.
We define an action as the execution of a single controller for $5$ seconds.

\vspace{-4mm}
\subsection{Results}

\begin{figure*}[ht!]
\includegraphics[width=.95\linewidth]{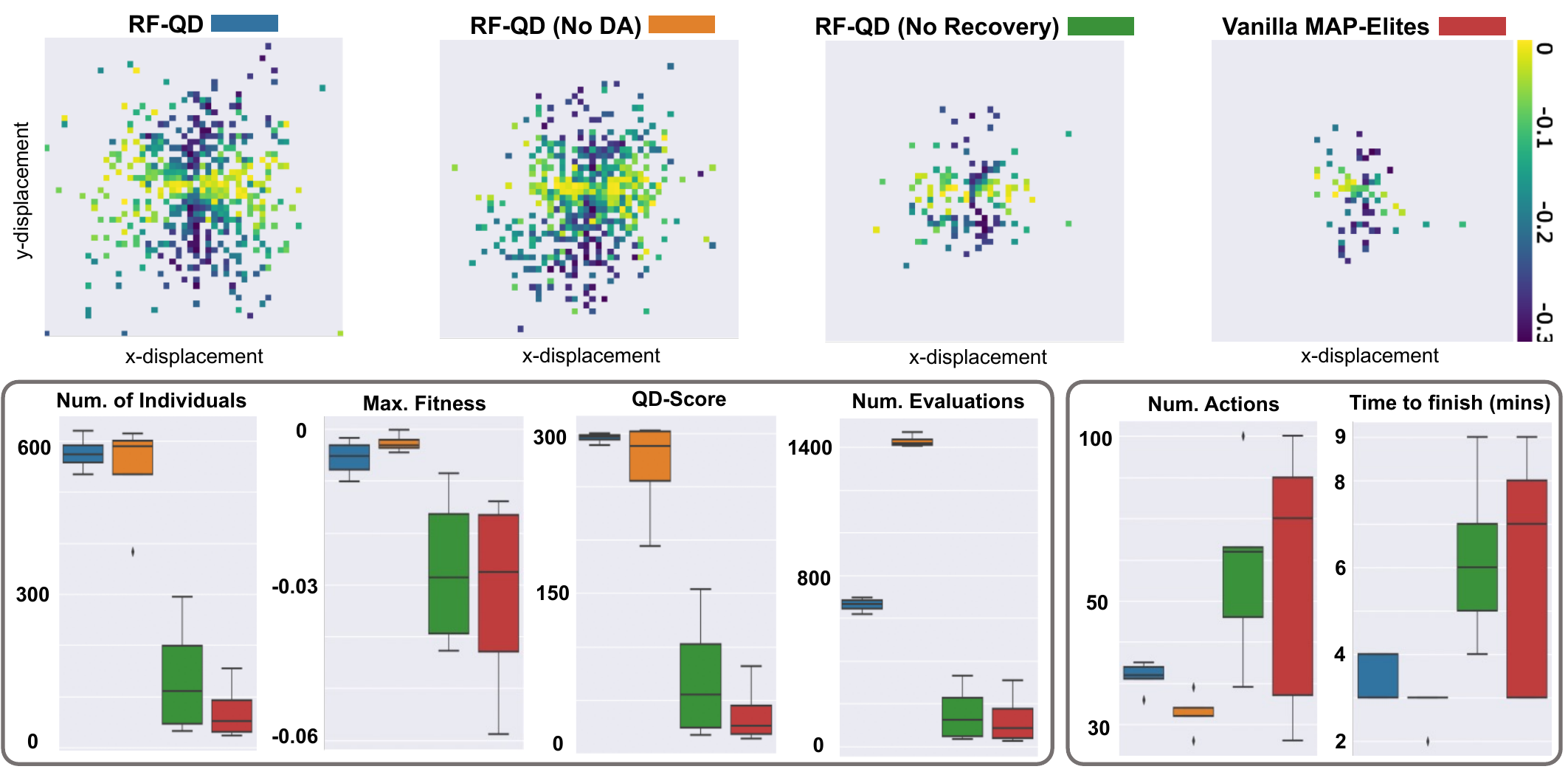} 
\vspace{-5pt}
\caption{
Top: without dynamics awareness (RF-QD no DA, orange), the archive tends to have more solutions around the origin compared to RF-QD (blue). For RF-QD no Recovery (green) and MAP-Elites (red), few individuals are found before the robot is outside the training zone. Bottom: Archive generation and navigation tasks have the best results for RF-QD and RF-QD no DA.}
\label{fig:best_archives}
\vspace{-3mm}
\end{figure*}

\textbf{Archive Generation}. Fig.~\ref{fig:best_archives} (top) shows the best archive for each algorithm. Algorithms that include recovery, RF-QD and RF-QD no DA, outperform those without it.
The coverage and maximum fitness values achieved by RF-QD and RF-QD no DA are equivalent and the best among the algorithms.
When no recovery function is available, the robot may leave the recovery zone, resulting in early termination and lower number of evaluations, Fig.~\ref{fig:best_archives} (bottom).
On average, the non-recovery algorithms run 100 evaluations before leaving the recovering zone. 
By contrast, the two algorithms with recovery run 600 and 1400 evaluations on average.
RF-QD no Recovery performs slightly better than MAP-Elites thanks to the behaviour selection mechanism.
This mechanism allows the algorithm to perform slightly more evaluations than MAP-Elites, but was still insufficient to avoid the robot to leave the recovery zone.

RF-QD is more consistent on QD score while using fewer evaluations (Fig.\ref{fig:best_archives}).
In the same time frame, RF-QD no DA executed more evaluations as it does not require time for training the model and predicting behaviours.
However, in physical training, requiring fewer evaluations is a desired outcome.
Also, asynchronous execution of controllers and training of the model can always be implemented to reduce this difference.
Moreover, fewer evaluations are desirable as it reduces the probability of damaging the robot. 

\textbf{Maze Navigation Task}. Fig.~\ref{fig:best_archives} also shows the results of each algorithm in the navigation task.
RF-QD and RF-QD no DA have the best results with fewer actions and shorter arrival times.
These results are related to the coverage, maximum fitness and QD score of the archives after training.
There is a slight increase in performance for the no DA variation.
This effect is related to the selection policy that prioritises individuals predicted to be safer and novel.


The results for the no-recovery versions show the relative worst performance. The lack of diversity and relatively low fitness results in the robot not having a proper set of actions to reach the goal. The robot was observed to get stuck in corners and hit obstacles.

\vspace{-8pt}
\section{Conclusions}

We trained a physical robot to find diverse walking solutions without using any physical simulations. 
After 2 hours of training, the robot could generate enough diverse solutions to navigate a maze.
Using a recovery function is necessary to keep the robot in the training regime. The data efficiency of our approach is improved by using the dynamics model to predict and select safe and novel controllers before testing them in the robot.
To the best of our knowledge, this is the first implementation of a QD algorithm directly on a physical robot without using simulations. A video showing the training and navigation tasks is available at https://youtu.be/BgGNvIsRh7Q

\vspace{-6pt}
\bibliographystyle{ACM-Reference-Format}
\bibliography{main}

\end{document}

%% file: notations.tex
\newcommand{\qd}{QD\xspace}
\newcommand{\bd}{BD\xspace}

\newcommand{\rfqd}{RF-QD\xspace}
\newcommand{\mapelites}{MAP-Elites\xspace}

%% file: main.bbl

\begin{thebibliography}{14}


\ifx \showCODEN    \undefined \def \showCODEN     #1{\unskip}     \fi
\ifx \showDOI      \undefined \def \showDOI       #1{#1}\fi
\ifx \showISBNx    \undefined \def \showISBNx     #1{\unskip}     \fi
\ifx \showISBNxiii \undefined \def \showISBNxiii  #1{\unskip}     \fi
\ifx \showISSN     \undefined \def \showISSN      #1{\unskip}     \fi
\ifx \showLCCN     \undefined \def \showLCCN      #1{\unskip}     \fi
\ifx \shownote     \undefined \def \shownote      #1{#1}          \fi
\ifx \showarticletitle \undefined \def \showarticletitle #1{#1}   \fi
\ifx \showURL      \undefined \def \showURL       {\relax}        \fi
\providecommand\bibfield[2]{#2}
\providecommand\bibinfo[2]{#2}
\providecommand\natexlab[1]{#1}
\providecommand\showeprint[2][]{arXiv:#2}

\bibitem[Akkaya et~al\mbox{.}(2019)]%
        {akkaya2019solving}
\bibfield{author}{\bibinfo{person}{Ilge Akkaya}, \bibinfo{person}{Marcin
  Andrychowicz}, \bibinfo{person}{Maciek Chociej}, \bibinfo{person}{Mateusz
  Litwin}, \bibinfo{person}{Bob McGrew}, \bibinfo{person}{Arthur Petron},
  \bibinfo{person}{Alex Paino}, \bibinfo{person}{Matthias Plappert},
  \bibinfo{person}{Glenn Powell}, \bibinfo{person}{Raphael Ribas},
  {et~al\mbox{.}}} \bibinfo{year}{2019}\natexlab{}.
\newblock \showarticletitle{Solving rubik's cube with a robot hand}.
\newblock \bibinfo{journal}{\emph{arXiv preprint arXiv:1910.07113}}
  (\bibinfo{year}{2019}).
\newblock


\bibitem[Allard et~al\mbox{.}(2022)]%
        {hbr}
\bibfield{author}{\bibinfo{person}{Maxime Allard},
  \bibinfo{person}{Sim\'{o}n~C. Smith}, \bibinfo{person}{Konstantinos
  Chatzilygeroudis}, {and} \bibinfo{person}{Antoine Cully}.}
  \bibinfo{year}{2022}\natexlab{}.
\newblock \showarticletitle{Hierarchical Quality-Diversity for Online Damage
  Recovery}. In \bibinfo{booktitle}{\emph{Proceedings of the Genetic and
  Evolutionary Computation Conference}} (Boston, Massachusetts)
  \emph{(\bibinfo{series}{GECCO '22})}. \bibinfo{publisher}{Association for
  Computing Machinery}, \bibinfo{address}{New York, NY, USA},
  \bibinfo{pages}{58–67}.
\newblock
\showISBNx{9781450392372}
\urldef\tempurl%
\url{https://doi.org/10.1145/3512290.3528751}
\showDOI{\tempurl}


\bibitem[Chatzilygeroudis et~al\mbox{.}(2016)]%
        {rte}
\bibfield{author}{\bibinfo{person}{Konstantinos~I. Chatzilygeroudis},
  \bibinfo{person}{Vassilis Vassiliades}, {and}
  \bibinfo{person}{Jean{-}Baptiste Mouret}.} \bibinfo{year}{2016}\natexlab{}.
\newblock \showarticletitle{Reset-free Trial-and-Error Learning for
  Data-Efficient Robot Damage Recovery}.
\newblock \bibinfo{journal}{\emph{CoRR}}  \bibinfo{volume}{abs/1610.04213}
  (\bibinfo{year}{2016}).
\newblock
\showeprint[arXiv]{1610.04213}
\urldef\tempurl%
\url{http://arxiv.org/abs/1610.04213}
\showURL{%
\tempurl}


\bibitem[Cully et~al\mbox{.}(2015)]%
        {nature}
\bibfield{author}{\bibinfo{person}{Antoine Cully}, \bibinfo{person}{Jeff
  Clune}, \bibinfo{person}{Danesh Tarapore}, {and}
  \bibinfo{person}{Jean-Baptiste Mouret}.} \bibinfo{year}{2015}\natexlab{}.
\newblock \showarticletitle{Robots that can adapt like animals}.
\newblock \bibinfo{journal}{\emph{Nature}} \bibinfo{volume}{521},
  \bibinfo{number}{7553} (\bibinfo{year}{2015}), \bibinfo{pages}{503--507}.
\newblock


\bibitem[Cully and Demiris(2018)]%
        {qdunifying}
\bibfield{author}{\bibinfo{person}{Antoine Cully} {and}
  \bibinfo{person}{Yiannis Demiris}.} \bibinfo{year}{2018}\natexlab{}.
\newblock \showarticletitle{Quality and diversity optimization: A unifying
  modular framework}.
\newblock \bibinfo{journal}{\emph{IEEE Transactions on Evolutionary
  Computation}} \bibinfo{volume}{22}, \bibinfo{number}{2}
  (\bibinfo{year}{2018}), \bibinfo{pages}{245--259}.
\newblock


\bibitem[Grillotti and Cully(2022)]%
        {aurora}
\bibfield{author}{\bibinfo{person}{Luca Grillotti} {and}
  \bibinfo{person}{Antoine Cully}.} \bibinfo{year}{2022}\natexlab{}.
\newblock \showarticletitle{Unsupervised Behavior Discovery With
  Quality-Diversity Optimization}.
\newblock \bibinfo{journal}{\emph{IEEE Transactions on Evolutionary
  Computation}} \bibinfo{volume}{26}, \bibinfo{number}{6}
  (\bibinfo{year}{2022}), \bibinfo{pages}{1539--1552}.
\newblock


\bibitem[Lim et~al\mbox{.}(2022a)]%
        {daqd}
\bibfield{author}{\bibinfo{person}{Bryan Lim}, \bibinfo{person}{Luca
  Grillotti}, \bibinfo{person}{Lorenzo Bernasconi}, {and}
  \bibinfo{person}{Antoine Cully}.} \bibinfo{year}{2022}\natexlab{a}.
\newblock \showarticletitle{Dynamics-aware quality-diversity for efficient
  learning of skill repertoires}. In \bibinfo{booktitle}{\emph{2022
  International Conference on Robotics and Automation (ICRA)}}. IEEE,
  \bibinfo{pages}{5360--5366}.
\newblock


\bibitem[Lim et~al\mbox{.}(2022b)]%
        {rfqd}
\bibfield{author}{\bibinfo{person}{Bryan Lim}, \bibinfo{person}{Alexander
  Reichenbach}, {and} \bibinfo{person}{Antoine Cully}.}
  \bibinfo{year}{2022}\natexlab{b}.
\newblock \showarticletitle{Learning to Walk Autonomously via Reset-Free
  Quality-Diversity}. In \bibinfo{booktitle}{\emph{Proceedings of the Genetic
  and Evolutionary Computation Conference}} (Boston, Massachusetts)
  \emph{(\bibinfo{series}{GECCO '22})}. \bibinfo{publisher}{Association for
  Computing Machinery}, \bibinfo{address}{New York, NY, USA},
  \bibinfo{pages}{86–94}.
\newblock
\showISBNx{9781450392372}
\urldef\tempurl%
\url{https://doi.org/10.1145/3512290.3528715}
\showDOI{\tempurl}


\bibitem[Mouret and Clune(2015)]%
        {mapelites}
\bibfield{author}{\bibinfo{person}{Jean-Baptiste Mouret} {and}
  \bibinfo{person}{Jeff Clune}.} \bibinfo{year}{2015}\natexlab{}.
\newblock \bibinfo{title}{Illuminating search spaces by mapping elites}.
\newblock
\newblock
\urldef\tempurl%
\url{https://doi.org/10.48550/ARXIV.1504.04909}
\showDOI{\tempurl}


\bibitem[Nagabandi et~al\mbox{.}(2020)]%
        {nagabandi}
\bibfield{author}{\bibinfo{person}{Anusha Nagabandi}, \bibinfo{person}{Kurt
  Konolige}, \bibinfo{person}{Sergey Levine}, {and} \bibinfo{person}{Vikash
  Kumar}.} \bibinfo{year}{2020}\natexlab{}.
\newblock \showarticletitle{Deep dynamics models for learning dexterous
  manipulation}. In \bibinfo{booktitle}{\emph{Conference on Robot Learning}}.
  PMLR, \bibinfo{pages}{1101--1112}.
\newblock


\bibitem[Vassiliades and Mouret(2018)]%
        {10.1145/3205455.3205602}
\bibfield{author}{\bibinfo{person}{Vassiiis Vassiliades} {and}
  \bibinfo{person}{Jean-Baptiste Mouret}.} \bibinfo{year}{2018}\natexlab{}.
\newblock \showarticletitle{Discovering the Elite Hypervolume by Leveraging
  Interspecies Correlation}. In \bibinfo{booktitle}{\emph{Proceedings of the
  Genetic and Evolutionary Computation Conference}} (Kyoto, Japan)
  \emph{(\bibinfo{series}{GECCO '18})}. \bibinfo{publisher}{Association for
  Computing Machinery}, \bibinfo{address}{New York, NY, USA},
  \bibinfo{pages}{149–156}.
\newblock
\showISBNx{9781450356183}
\urldef\tempurl%
\url{https://doi.org/10.1145/3205455.3205602}
\showDOI{\tempurl}


\bibitem[Wu et~al\mbox{.}(2022)]%
        {daydreamer}
\bibfield{author}{\bibinfo{person}{Philipp Wu}, \bibinfo{person}{Alejandro
  Escontrela}, \bibinfo{person}{Danijar Hafner}, \bibinfo{person}{Ken
  Goldberg}, {and} \bibinfo{person}{Pieter Abbeel}.}
  \bibinfo{year}{2022}\natexlab{}.
\newblock \bibinfo{title}{DayDreamer: World Models for Physical Robot
  Learning}.
\newblock
\newblock
\urldef\tempurl%
\url{https://doi.org/10.48550/ARXIV.2206.14176}
\showDOI{\tempurl}


\bibitem[Yang et~al\mbox{.}(2020)]%
        {dataefificentrl}
\bibfield{author}{\bibinfo{person}{Yuxiang Yang}, \bibinfo{person}{Ken
  Caluwaerts}, \bibinfo{person}{Atil Iscen}, \bibinfo{person}{Tingnan Zhang},
  \bibinfo{person}{Jie Tan}, {and} \bibinfo{person}{Vikas Sindhwani}.}
  \bibinfo{year}{2020}\natexlab{}.
\newblock \showarticletitle{Data efficient reinforcement learning for legged
  robots}. In \bibinfo{booktitle}{\emph{Conference on Robot Learning}}. PMLR,
  \bibinfo{pages}{1--10}.
\newblock


\bibitem[Zhang et~al\mbox{.}(2019)]%
        {solar}
\bibfield{author}{\bibinfo{person}{Marvin Zhang}, \bibinfo{person}{Sharad
  Vikram}, \bibinfo{person}{Laura Smith}, \bibinfo{person}{Pieter Abbeel},
  \bibinfo{person}{Matthew Johnson}, {and} \bibinfo{person}{Sergey Levine}.}
  \bibinfo{year}{2019}\natexlab{}.
\newblock \showarticletitle{Solar: Deep structured representations for
  model-based reinforcement learning}. In
  \bibinfo{booktitle}{\emph{International conference on machine learning}}.
  PMLR, \bibinfo{pages}{7444--7453}.
\newblock


\end{thebibliography}
